\documentclass[sigplan,screen]{acmart}
\AtBeginDocument{%
  \providecommand\BibTeX{{%
    \normalfont B\kern-0.5em{\scshape i\kern-0.25em b}\kern-0.8em\TeX}}}

\setcopyright{acmcopyright}
\copyrightyear{2018}
\acmYear{2018}
\acmDOI{XXXXXXX.XXXXXXX}

\acmConference[Conference acronym 'XX]{Make sure to enter the correct
  conference title from your rights confirmation emai}{June 03--05,
  2018}{Woodstock, NY}
\acmPrice{15.00}
\acmISBN{978-1-4503-XXXX-X/18/06}

\begin{document}

\title{Development of a Neural Network-Based Mathematical Operation Protocol
for Embedded Hexadecimal Digits Using Neural Architecture Search (NAS)}

\author{Victor Robila}
\email{victor.robila@gmail.com}
\authornotemark[1]
\affiliation{%
  \institution{Hunter College High School}
  \city{New York}
  \state{New York}
  \country{USA}
  \postcode{11432}
}

\author{Kexin Pei}
\email{kpei@cs.columbia.edu}
\affiliation{%
  \institution{Columbia University}
  \city{New York}
  \country{New York}
  \country{USA}
}

\author{Junfeng Yang}
\email{junfeng@cs.columbia.edu}
\affiliation{%
 \institution{Columbia University}
  \city{New York}
  \country{New York}
  \country{USA}
 }

This work may be submitted to the IEEE for possible publication. Copyright may be transferred without notice, after which this version may no longer be accessible.

\renewcommand{\shortauthors}{Robila, et al.}

\begin{abstract}
  It is beneficial to develop an efficient machine-learning based method for addition using embedded hexadecimal digits. Through a comparison between human-developed machine learning model and models sampled through Neural Architecture Search (NAS) we determine an efficient approach to solve this problem with a final testing loss of 0.2937 for a human-developed model. 
\end{abstract}

\keywords{datasets, neural networks, machine learning, hexadecimal}


\maketitle

\section{Introduction}
Mathematical operations, mainly addition, subtraction, division, and
multiplication remain at the very foundation of all mathematical
concepts \cite{whitehead2017introduction}. Neural Networks have also been used extensively in
mathematical applications, with most approaches focusing on problems
from competitions like AMC 10, AMC 12, and AIME or university-level math
problems \cite{voevoda2017synthesis}\cite{drori2022neural}\cite{hendrycks2021measuring}.~
\\
\\
\vspace{.1cm}\noindent\textbf{Our approach.}
We develop a machine learning based method to predict the results of
these four operations when given PyTorch embeddings of hexadecimal
digits \cite{embedding, imambi2021pytorch}. A variety of model types are tested and ranked
based on a loss metric. Three different humans developed PyTorch neural
networks are first described, trained, tested, and evaluated. These
human-developed networks all have different architectures: one uses a
Fully Connected neural network, one uses an LSTM layer, and the final
uses self-attention. Following the creation of these networks, Neural
Architecture Search (NAS) has been performed on the base linear layer
neural network using Microsoft's Neural Network Intelligence Package
\cite{elsken2019neural, nni}. The human-developed neural networks and searched
networks were then compared to each other.~

For this project three different simulated datasets were used in order
to simultaneously observe if the number of hex digits that are
represented had an effect on the model's final loss value. This test was
inspired by Nogueira et al.'s findings which indicated that the way a
number is represented can affect accuracy in mathematical operations
\cite{nogueira2021investigating}.

These experiments were done as part of a larger project that focused on
machine learning-based interpretations of human code and is a way to
improve efficiency in that program by acting as a subroutine that
removes the need for the larger model to recognize and compute
mathematical operations.

An outline of this write-up is as follows: Section 2. describes related
work in the field of machine learning addition problems, Section 3
describes our methods, Section 4. describes results, and Section 5.
communicates our conclusions and future work. Section 3. includes
information on all models created as part of this project. These models
can be used when normal numerical operations are unavailable, such as
when using embeddings that lose information about the original numbers.

\section{Related Work}

\vspace{.1cm}\noindent\textbf{Neural Networks in mathematics}
Neural networks have been previously used to solve mathematical problems
from competitions such as those provided using the MATH dataset proposed
by Hendrycks et al. (2021) \cite{hendrycks2021measuring}. With the MATH dataset, problems are
taken from competitions such as AMC 10, AMC 12, AIME, and others and are
given scores. Following the creation of the dataset, several models such
as GPT-3 and GPT-2 were tested and ranked on their problem-solving
ability. Theephoowiang et al. (2022) use neural networks to estimate the
difficulty of mathematics problems \cite{theephoowiang2022difficulty}.

\vspace{.1cm}\noindent\textbf{Transformers in Mathematics}
It has been shown that Transformers can perform difficult calculations
similarly to most calculators or computer systems. Amini et al. (2019),
and Ling et al. (2017) use plug and chug mathematics problems that are
multiple choice to observe sequence-to-program generation
\cite{amini2019mathqa}\cite{ling2017program}. Saxton et al. (2019) also developed the DeepMind
Mathematics dataset that includes some problems (mainly addition) that
are similar to those solved in this project \cite{saxton2019analysing}. Henighan et al.
(2020) showed that most problems in the DeepMind Mathematics dataset
were solvable with large transformers \cite{henighan2020scaling}. Lample and Charton also
used Transformers to solve symbolic integration problems and reached
more than 95\% accuracy \cite{lample2019deep}.

\vspace{.1cm}\noindent\textbf{Number representations}
The representation of numbers has also been found to affect results when
entered into a neural network. Nogueira et al. (2021) found that
introducing position tokens (e.g., ``2 10e1 2) can lead to improved
performance \cite{nogueira2021investigating}.

\vspace{.1cm}\noindent\textbf{Neural Architecture Search and Hyperparamenter Optimization}
Neural architecture search has been a growing interest in recent times
\cite{stanley2009hypercube}\cite{wierstra2005modeling}. Hyperparameter optimization is especially important
for cases where the actual network architecture in terms of layers is
relatively optimized and is increasingly being used
\cite{snoek2012practical}\cite{bergstra2012random} \cite{bergstra2011algorithms}. Network search protocols have also been used
to create new models \cite{floreano2008neuroevolution}\cite{stanley2009hypercube}\cite{wierstra2005modeling}. There have been several
different neural architecture search algorithms created for
PyTorch-based neural networks such as Microsoft's Neural Network
Intelligence, Auto-Pytorch, and Efficient Neural Architecture Search
(ENAS) \cite{nni}\cite{automl}\cite{carpedm20}.

\section{Methods}
\subsection{Overview of Software Components}

The software components used in the creation of the machine learning
models are Python, PyTorch, torchviz and Neural Network Intelligence.~

Python is an object-oriented high-level programming language and is one
of the most popular programming languages. Many machine learning
libraries are based in Python due to its ease of use, and libraries. For
this project we used Python version 3.8.10 \cite{python.org}. PyTorch is an
open-source machine learning framework that has many applications in
computer vision and natural language processing. We chose to use PyTorch
due to its simplicity and since the larger machine learning project that
this subroutine is part of uses PyTorch. Pytorch was used to program all
neural networks in this experiment. We used PyTorch version 1.10.2
\cite{pytorch}.~

Neural Network Intelligence (NNI) is a toolkit that is used to run
automated machine learning (AutoML)  \begin{figure}[h]
  \centering
  \includegraphics[width=\linewidth]{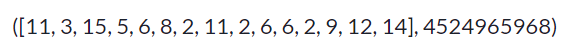}
  \caption{Example of an n-gram}
  \Description{List of values that compose an n-gram}
  \label{fig:1}
\end{figure} experiments
created by Microsoft.
Neural Network Intelligence was chosen for its Neural Architecture
Search capability. We used NNI version 2.7 \cite{nni}. Torchviz is a
package used to ``visualize PyTorch execution graphs.'' Torchviz version
x.xx was used to develop visualizations for our machine learning models
\cite{pypi}.

The Visual Studio Code Integrated Development Environment (IDE) was used
to write and run all the code described in this publication \cite{microsoft_2021}.

\subsection{Data Format and Generation}

Our primary goal was to predict the result of four arithmetic
operations: addition, subtraction, multiplication, and division. All
numerical data were originally in the form of 4-digit hexadecimal
integers. An example of a conversion table for hexadecimal and decimal
for numbers up to 256 can be found in Figure~\ref{fig:1}.

We created one dataset for each operation that resulted in a total of 4
datasets. Each dataset had 500,000 pairs and 500,000 results that
depended on which operation each dataset was based on. In training and
testing, a selection of these 500,000 values was taken and inputted into
the model in the form of n-grams. All of these values were embedded
using PyTorch embeddings of dimension 8 with data values. The final data
values submitted to all the models have 16 integer values that make up the embedded 4-digit hexadecimal digits and the real sum in
base 10. Figure 2. displays an example of two of these n-grams.

As discussed earlier, a paper found that the way numbers are represented
influences accuracy \cite{nogueira2021investigating}. Therefore, as a separate experiment we
decided to change the number of hex digits represented to see if this
had any effect on the loss. We did experiments with the full data, 3 hex
digits, and 2 hex digits.

\subsection{Human-Developed Machine Learning Models}

We developed 12 network architectures for use with our embedded hex
mathematical operation problems. We had 4 versions (one for each
operation) of three different architectures. Each network was trained on
each of the three data representations and the evaluation metric was the
loss produced by the network after the final epoch training.

\subsubsection{Human Generated Model Definitions}

\begin{figure}[h]
  \centering
  \includegraphics[width=\linewidth]{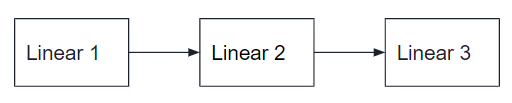}
  \caption{Visualization of Fully Connected Linear Neural Network}
  \Description{Architecture of a Fully-Connected Linear Neural Network}
  \label{fig:2}
\end{figure}

Our first machine learning model was a model with 3 PyTorch linear
layers and served as the base model for our Neural Network Intelligence
ValueChoice model.  Figure~\ref{fig:2} shows the architecture of this first model
.\begin{figure}[h]
  \centering
  \includegraphics[width=\linewidth]{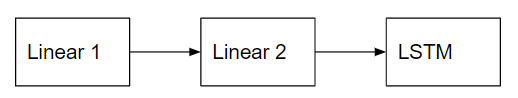}
  \caption{Visualization of LSTM Neural Network}
  \Description{Architecture of a LSTM Neural Network}
  \label{fig:3}
\end{figure}\begin{figure}[h]
  \centering
  \includegraphics[width=\linewidth]{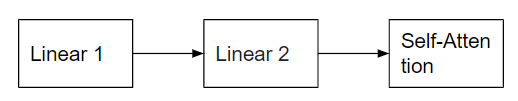}
  \caption{Visualization of Self Attention Neural Network}
  \Description{Architecture of a Self Attention Neural Network}
  \label{fig:4}
\end{figure}
Our next model used 
a Long Short-Term Memory layer and our final model
used a self-attention layer.  Figure~\ref{fig:3} shows the architecture for the self-attention neural network and Figure~\ref{fig:4} displays the network architecture
for the LSTM. All three models were trained three times using different
kinds of data encoding for each trial as described in section 3.2.

\subsection{Neural Architecture Search (NAS) Machine Learning Models}

\subsubsection{ValueChoice Model}

We used Neural Network Intelligence's ValueChoice feature using our fully connected neural network in order to see if it would be possible to improve the performance of the model by changing the number of connections between neurons. We gave Neural Network Intelligence 4 possibilities for the number of connections: 16, 32, 64, and 128. Our original fully connected neural network used 64 for this parameter. The final goal of this experiment was to produce one network that was the best out of the parameters provided to Neural Network Intelligence. We used "Random" for our search strategy and allowed Neural Network Intelligence to go up to 5 epochs.

\subsubsection{LayerChoice Model}

We also used Neural Network Intelligence's LayerChoice feature to see the effects of using Neural Architecture Search to change the network architecture. Similar to our ValueChoice model we used our original fully connected neural network and allowed Neural Network Intelligence to develop networks through that. We gave Neural Network Intelligence the choice between an identity layer and a linear layer and allowed it to find the best one. We used "Random" for our search strategy and allowed Neural Network Intelligence to go up to 5 epochs.\begin{figure}[h]
  \centering
  \includegraphics[width=\linewidth]{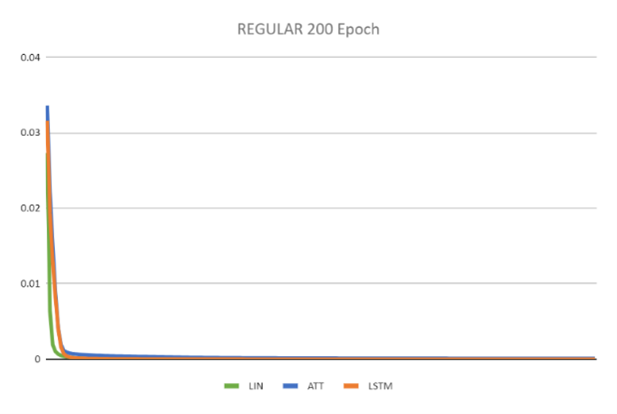}
  \caption{Comparison between human-generated machine learning models for 4-digit hexadecimal values for 200 epochs}
  \Description{Graph of model performance}
  \label{fig:5}
\end{figure}\begin{figure}[h]
  \centering
  \includegraphics[width=\linewidth]{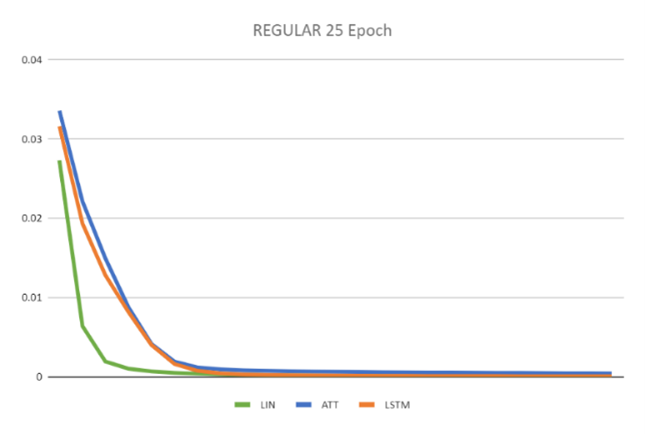}
  \caption{Comparison between human-generated machine learning models for 4-digit hexadecimal values for 25  epochs}
  \Description{Graph of model performance}
  \label{fig:6}
\end{figure}\begin{figure}[h]
  \centering
  \includegraphics[width=\linewidth]{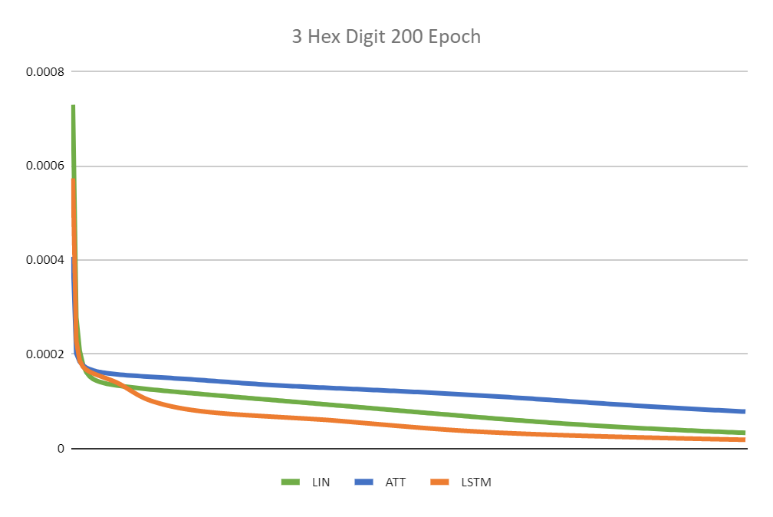}
  \caption{Comparison between human-generated machine learning models for 3-digit hexadecimal values for 200 epochs}
  \Description{Graph of model performance}
  \label{fig:7}
\end{figure}\begin{figure}[h]
  \centering
  \includegraphics[width=\linewidth]{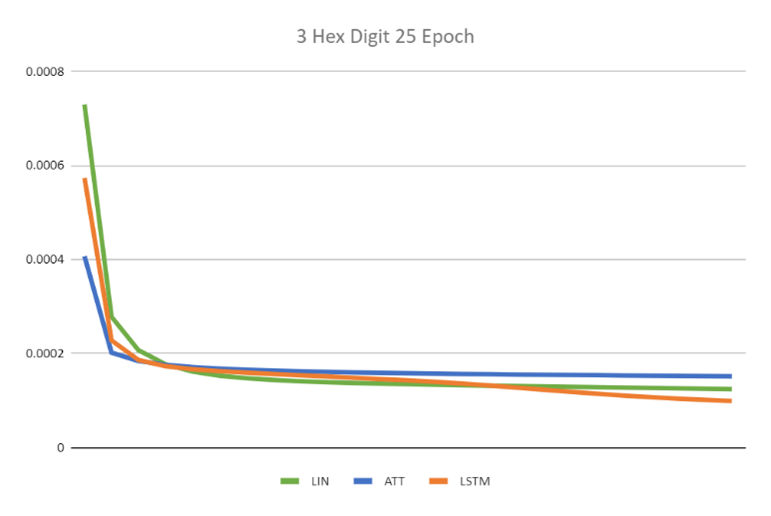}
  \caption{Comparison between human-generated machine learning models for 3-digit hexadecimal values for 25 epochs}
  \Description{Graph of model performance}
  \label{fig:8}
\end{figure}\begin{figure}[h]
  \centering
  \includegraphics[width=\linewidth]{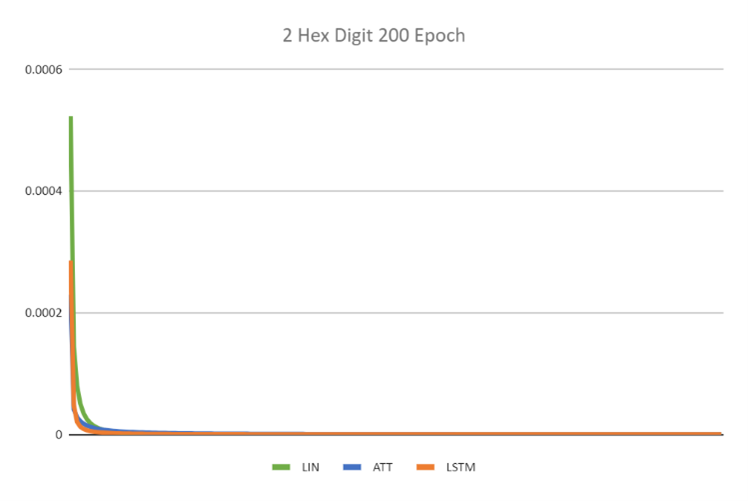}
  \caption{Comparison between human-generated machine learning models for 2-digit hexadecimal values for 200 epochs}
  \Description{Graph of model performance}
  \label{fig:9}
\end{figure}\begin{figure}[h]
  \centering
  \includegraphics[width=\linewidth]{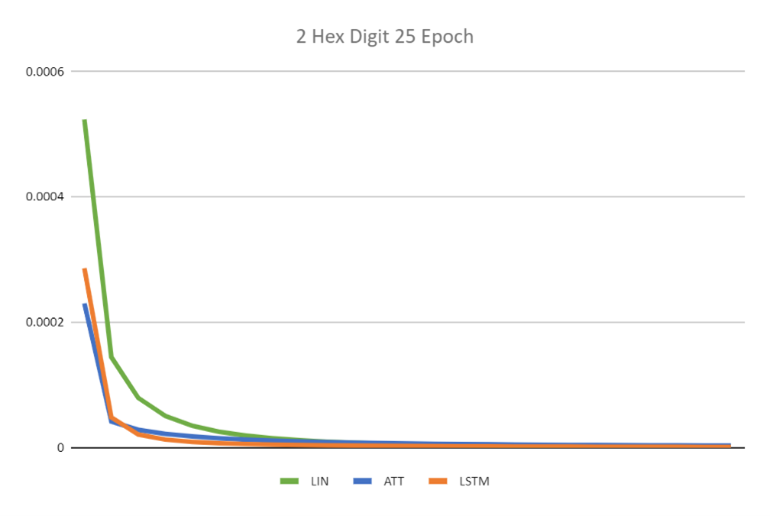}
  \caption{Comparison between human-generated machine learning models for 2-digit hexadecimal values for 25 epochs}
  \Description{Graph of model performance}
  \label{fig:10}
\end{figure}\begin{table*}
  \caption{Final loss after 200 epochs and on testing data for all human-generated networks}
  \label{tab:1}
\begin{tabular}{|l|l|l|l|}
\hline
Model                     & Loss after 200 Epochs & Loss for Testing Data   \\ \hline
Fully Connected, 4 Digits & 0.0000005605330506    & \textbf{0.0000005576385969}     \\ \hline
Self-Attention, 4 Digits  & 0.0001048754811       & 0.000111305436          \\ \hline
LSTM, 4 Digits            & 0.000005764168952     & 0.000005698243726       \\ \hline
Fully Connected, 3 Digits & 0.00003351484126      & 0.2713421421            \\ \hline
Self-Attention, 3 Digits  & 0.00007848687369      &\textbf{0.2675585822 }            \\ \hline
LSTM, 3 Digits            & 0.00001873845527      & 0.2765113088            \\ \hline
Fully Connected, 2 Digits & 0.0000005991071621    & 0.2936991373            \\ \hline
Self-Attention, 2 Digits  & 0.0000007127712775    & 0.2945040227            \\ \hline
LSTM, 2 Digits            & 0.0000006743473424    & \textbf{0.2730684676 }           \\ \hline
\end{tabular}
\end{table*}
 Since we wanted to see how well the models extrapolated on larger data, we trained them on 2 digit hex numbers and tested them on 4 digit hex numbers.

\section{Results}

\subsection{Comparison between human-generated machine learning models}

The models for every operation (addition, subtraction, multiplication,
and division) were based on addition (multiplication and division were
done using logs), and thus this comparison will focus on addition
operation and the three data sizes (4-digit hex, 3-digit hex, and
2-digit hex).

Each neural network seemed to have benefits in different situations and the
comparisons can be seen in Figures ~\ref{tab:1}-~\ref{fig:5} and Table~\ref{fig:10}. Figures~\ref{fig:5},~\ref{fig:7}, and ~\ref{fig:9}
show the loss as the models were training for 200 epochs and figures ~\ref{fig:6},~\ref{fig:7}, and ~\ref{fig:8} show the loss as the models train on the same data for 25
epochs. Table~\ref{tab:1} shows the final losses for the networks on the training
dataset and on testing data. As seen from Figures~\ref{fig:5} and ~\ref{fig:6}, and Table 1,
the Fully Connected neural network fits the fastest to the training data
and has the lowest loss on both the testing and training data after 200
epochs. When the data are reduced to only include embedded versions of 3
hexadecimal digits in Figures~\ref{fig:7}, and ~\ref{fig:8}, the Self-Attention neural
network fits fastest at first but is quickly\begin{table}
  \caption{Loss on testing data after 200 epochs for Neural Architecture Search sourced neural network and human-generated fully connected neural network}
  \label{tab:2}
\begin{tabular}{|l|l|}
\hline
NAS          & Human                 \\ \hline
0.2940571394 & \textbf{0.2936991373} \\ \hline
\end{tabular}
\end{table} surpassed by the other two networks. The LSTM network comes out to be the best on the training
data around the 15\textsuperscript{th} epoch. However, when the loss on
the testing data is viewed, the self-attention network has the lowest
loss. It is unclear why this occurs and requires further testing. All
the networks were similar on the 2-digit hexadecimal values in Figures ~\ref{fig:9} and ~\ref{fig:10} but the Fully Connected network seems to be the slowest to fit to
the training data. After running the models on the testing data, the
LSTM network had the lowest loss.

\subsection{Comparison of Human-Generated Networks and NAS-Searched Networks}

After running Neural Network Intelligence using ValueChoice, we found that the best network in the number of epochs we gave it used 32 connections. Through looking at the raw data, we were in fact able to determine that it did have the lowest loss within the epoch range but noticed that it was increasing after each epoch and that 16 and 128 seemed to be converging better. 

Using LayerChoice, we found that the architecture with only two linear layers and one identity layer was the best. Figures 11 and 12 show the loss for the Neural Architecture Searched model and the original human-generated model and from the graphs it is obvious that the NAS network converges a lot faster and fits to the data well. However, when testing on the testing data after training for 200 epochs we generated the results in Table~\ref{tab:2}. The human-generated loss is slightly less but this may be a result of over-fitting in the NAS network.

\section{Conclusions and Future Work}

All networks in the project worked and generated results that could be
useful given a need for embedded hexadecimal digits. The NAS-sourced
models were comparable to the human-generated models and did converge faster in the LayerChoice examples. There were some unexpected results in that the human-generated network was better than the NAS sourced network in the testing stage of the LayerChoice experiment. There are several directions for future research in this area, including searching a wider model space in the NAS section of our research and possibly developing a larger dataset if given more processing power.

\bibliographystyle{ACM-Reference-Format}
\bibliography{sample-base}


\begin{thebibliography}{27}


\ifx \showCODEN    \undefined \def \showCODEN     #1{\unskip}     \fi
\ifx \showDOI      \undefined \def \showDOI       #1{#1}\fi
\ifx \showISBNx    \undefined \def \showISBNx     #1{\unskip}     \fi
\ifx \showISBNxiii \undefined \def \showISBNxiii  #1{\unskip}     \fi
\ifx \showISSN     \undefined \def \showISSN      #1{\unskip}     \fi
\ifx \showLCCN     \undefined \def \showLCCN      #1{\unskip}     \fi
\ifx \shownote     \undefined \def \shownote      #1{#1}          \fi
\ifx \showarticletitle \undefined \def \showarticletitle #1{#1}   \fi
\ifx \showURL      \undefined \def \showURL       {\relax}        \fi
\providecommand\bibfield[2]{#2}
\providecommand\bibinfo[2]{#2}
\providecommand\natexlab[1]{#1}
\providecommand\showeprint[2][]{arXiv:#2}

\bibitem[aut(2022)]%
        {automl}
 \bibinfo{year}{2022}\natexlab{}.
\newblock \bibinfo{title}{Auto-pytorch}.
\newblock
\newblock
\urldef\tempurl%
\url{https://www.automl.org/automl/autopytorch/}
\showURL{%
\tempurl}


\bibitem[emb(2022)]%
        {embedding}
 \bibinfo{year}{2022}\natexlab{}.
\newblock \bibinfo{title}{Embedding}.
\newblock
\newblock
\urldef\tempurl%
\url{https://pytorch.org/docs/stable/generated/torch.nn.Embedding.html}
\showURL{%
\tempurl}


\bibitem[nni(2022)]%
        {nni}
 \bibinfo{year}{2022}\natexlab{}.
\newblock \bibinfo{title}{NNI documentation}.
\newblock
\newblock
\urldef\tempurl%
\url{https://nni.readthedocs.io/en/stable/}
\showURL{%
\tempurl}


\bibitem[pyt(2022a)]%
        {python.org}
 \bibinfo{year}{2022}\natexlab{a}.
\newblock \bibinfo{title}{Python release python 3.8.10}.
\newblock
\newblock
\urldef\tempurl%
\url{https://www.python.org/downloads/release/python-3810/}
\showURL{%
\tempurl}


\bibitem[pyt(2022b)]%
        {pytorch}
 \bibinfo{year}{2022}\natexlab{b}.
\newblock \bibinfo{title}{Pytorch}.
\newblock
\newblock
\urldef\tempurl%
\url{https://pytorch.org/}
\showURL{%
\tempurl}


\bibitem[pyp(2022)]%
        {pypi}
 \bibinfo{year}{2022}\natexlab{}.
\newblock \bibinfo{title}{Torchviz}.
\newblock
\newblock
\urldef\tempurl%
\url{https://pypi.org/project/torchviz/}
\showURL{%
\tempurl}


\bibitem[Amini et~al\mbox{.}(2019)]%
        {amini2019mathqa}
\bibfield{author}{\bibinfo{person}{Aida Amini}, \bibinfo{person}{Saadia
  Gabriel}, \bibinfo{person}{Peter Lin}, \bibinfo{person}{Rik
  Koncel-Kedziorski}, \bibinfo{person}{Yejin Choi}, {and}
  \bibinfo{person}{Hannaneh Hajishirzi}.} \bibinfo{year}{2019}\natexlab{}.
\newblock \showarticletitle{MathQA: Towards interpretable math word problem
  solving with operation-based formalisms}.
\newblock \bibinfo{journal}{\emph{arXiv preprint arXiv:1905.13319}}
  (\bibinfo{year}{2019}).
\newblock


\bibitem[Bergstra et~al\mbox{.}(2011)]%
        {bergstra2011algorithms}
\bibfield{author}{\bibinfo{person}{James Bergstra}, \bibinfo{person}{R{\'e}mi
  Bardenet}, \bibinfo{person}{Yoshua Bengio}, {and} \bibinfo{person}{Bal{\'a}zs
  K{\'e}gl}.} \bibinfo{year}{2011}\natexlab{}.
\newblock \showarticletitle{Algorithms for hyper-parameter optimization}.
\newblock \bibinfo{journal}{\emph{Advances in neural information processing
  systems}}  \bibinfo{volume}{24} (\bibinfo{year}{2011}).
\newblock


\bibitem[Bergstra and Bengio(2012)]%
        {bergstra2012random}
\bibfield{author}{\bibinfo{person}{James Bergstra} {and}
  \bibinfo{person}{Yoshua Bengio}.} \bibinfo{year}{2012}\natexlab{}.
\newblock \showarticletitle{Random search for hyper-parameter optimization.}
\newblock \bibinfo{journal}{\emph{Journal of machine learning research}}
  \bibinfo{volume}{13}, \bibinfo{number}{2} (\bibinfo{year}{2012}).
\newblock


\bibitem[carpedm20(2022)]%
        {carpedm20}
\bibfield{author}{\bibinfo{person}{carpedm20}.}
  \bibinfo{year}{2022}\natexlab{}.
\newblock \bibinfo{title}{Carpedm20/Enas-pytorch: Pytorch implementation of
  "Efficient neural architecture search via parameters sharing"}.
\newblock
\newblock
\urldef\tempurl%
\url{https://github.com/carpedm20/ENAS-pytorch}
\showURL{%
\tempurl}


\bibitem[Drori et~al\mbox{.}(2022)]%
        {drori2022neural}
\bibfield{author}{\bibinfo{person}{Iddo Drori}, \bibinfo{person}{Sarah Zhang},
  \bibinfo{person}{Reece Shuttleworth}, \bibinfo{person}{Leonard Tang},
  \bibinfo{person}{Albert Lu}, \bibinfo{person}{Elizabeth Ke},
  \bibinfo{person}{Kevin Liu}, \bibinfo{person}{Linda Chen},
  \bibinfo{person}{Sunny Tran}, \bibinfo{person}{Newman Cheng},
  {et~al\mbox{.}}} \bibinfo{year}{2022}\natexlab{}.
\newblock \showarticletitle{A neural network solves, explains, and generates
  university math problems by program synthesis and few-shot learning at human
  level}.
\newblock \bibinfo{journal}{\emph{Proceedings of the National Academy of
  Sciences}} \bibinfo{volume}{119}, \bibinfo{number}{32}
  (\bibinfo{year}{2022}), \bibinfo{pages}{e2123433119}.
\newblock


\bibitem[Elsken et~al\mbox{.}(2019)]%
        {elsken2019neural}
\bibfield{author}{\bibinfo{person}{Thomas Elsken}, \bibinfo{person}{Jan~Hendrik
  Metzen}, {and} \bibinfo{person}{Frank Hutter}.}
  \bibinfo{year}{2019}\natexlab{}.
\newblock \showarticletitle{Neural architecture search: A survey}.
\newblock \bibinfo{journal}{\emph{The Journal of Machine Learning Research}}
  \bibinfo{volume}{20}, \bibinfo{number}{1} (\bibinfo{year}{2019}),
  \bibinfo{pages}{1997--2017}.
\newblock


\bibitem[Floreano et~al\mbox{.}(2008)]%
        {floreano2008neuroevolution}
\bibfield{author}{\bibinfo{person}{Dario Floreano}, \bibinfo{person}{Peter
  D{\"u}rr}, {and} \bibinfo{person}{Claudio Mattiussi}.}
  \bibinfo{year}{2008}\natexlab{}.
\newblock \showarticletitle{Neuroevolution: from architectures to learning}.
\newblock \bibinfo{journal}{\emph{Evolutionary intelligence}}
  \bibinfo{volume}{1}, \bibinfo{number}{1} (\bibinfo{year}{2008}),
  \bibinfo{pages}{47--62}.
\newblock


\bibitem[Hendrycks et~al\mbox{.}(2021)]%
        {hendrycks2021measuring}
\bibfield{author}{\bibinfo{person}{Dan Hendrycks}, \bibinfo{person}{Collin
  Burns}, \bibinfo{person}{Saurav Kadavath}, \bibinfo{person}{Akul Arora},
  \bibinfo{person}{Steven Basart}, \bibinfo{person}{Eric Tang},
  \bibinfo{person}{Dawn Song}, {and} \bibinfo{person}{Jacob Steinhardt}.}
  \bibinfo{year}{2021}\natexlab{}.
\newblock \showarticletitle{Measuring mathematical problem solving with the
  math dataset}.
\newblock \bibinfo{journal}{\emph{arXiv preprint arXiv:2103.03874}}
  (\bibinfo{year}{2021}).
\newblock


\bibitem[Henighan et~al\mbox{.}(2020)]%
        {henighan2020scaling}
\bibfield{author}{\bibinfo{person}{Tom Henighan}, \bibinfo{person}{Jared
  Kaplan}, \bibinfo{person}{Mor Katz}, \bibinfo{person}{Mark Chen},
  \bibinfo{person}{Christopher Hesse}, \bibinfo{person}{Jacob Jackson},
  \bibinfo{person}{Heewoo Jun}, \bibinfo{person}{Tom~B Brown},
  \bibinfo{person}{Prafulla Dhariwal}, \bibinfo{person}{Scott Gray},
  {et~al\mbox{.}}} \bibinfo{year}{2020}\natexlab{}.
\newblock \showarticletitle{Scaling laws for autoregressive generative
  modeling}.
\newblock \bibinfo{journal}{\emph{arXiv preprint arXiv:2010.14701}}
  (\bibinfo{year}{2020}).
\newblock


\bibitem[Imambi et~al\mbox{.}(2021)]%
        {imambi2021pytorch}
\bibfield{author}{\bibinfo{person}{Sagar Imambi}, \bibinfo{person}{Kolla~Bhanu
  Prakash}, {and} \bibinfo{person}{GR Kanagachidambaresan}.}
  \bibinfo{year}{2021}\natexlab{}.
\newblock \showarticletitle{PyTorch}.
\newblock In \bibinfo{booktitle}{\emph{Programming with TensorFlow}}.
  \bibinfo{publisher}{Springer}, \bibinfo{pages}{87--104}.
\newblock


\bibitem[Lample and Charton(2019)]%
        {lample2019deep}
\bibfield{author}{\bibinfo{person}{Guillaume Lample} {and}
  \bibinfo{person}{Fran{\c{c}}ois Charton}.} \bibinfo{year}{2019}\natexlab{}.
\newblock \showarticletitle{Deep learning for symbolic mathematics}.
\newblock \bibinfo{journal}{\emph{arXiv preprint arXiv:1912.01412}}
  (\bibinfo{year}{2019}).
\newblock


\bibitem[Ling et~al\mbox{.}(2017)]%
        {ling2017program}
\bibfield{author}{\bibinfo{person}{Wang Ling}, \bibinfo{person}{Dani Yogatama},
  \bibinfo{person}{Chris Dyer}, {and} \bibinfo{person}{Phil Blunsom}.}
  \bibinfo{year}{2017}\natexlab{}.
\newblock \showarticletitle{Program induction by rationale generation: Learning
  to solve and explain algebraic word problems}.
\newblock \bibinfo{journal}{\emph{arXiv preprint arXiv:1705.04146}}
  (\bibinfo{year}{2017}).
\newblock


\bibitem[Microsoft(2021)]%
        {microsoft_2021}
\bibfield{author}{\bibinfo{person}{Microsoft}.}
  \bibinfo{year}{2021}\natexlab{}.
\newblock \bibinfo{title}{Visual studio code - code editing. redefined}.
\newblock
\newblock
\urldef\tempurl%
\url{https://code.visualstudio.com/}
\showURL{%
\tempurl}


\bibitem[Nogueira et~al\mbox{.}(2021)]%
        {nogueira2021investigating}
\bibfield{author}{\bibinfo{person}{Rodrigo Nogueira}, \bibinfo{person}{Zhiying
  Jiang}, {and} \bibinfo{person}{Jimmy Lin}.} \bibinfo{year}{2021}\natexlab{}.
\newblock \showarticletitle{Investigating the limitations of transformers with
  simple arithmetic tasks}.
\newblock \bibinfo{journal}{\emph{arXiv preprint arXiv:2102.13019}}
  (\bibinfo{year}{2021}).
\newblock


\bibitem[Saxton et~al\mbox{.}(2019)]%
        {saxton2019analysing}
\bibfield{author}{\bibinfo{person}{David Saxton}, \bibinfo{person}{Edward
  Grefenstette}, \bibinfo{person}{Felix Hill}, {and} \bibinfo{person}{Pushmeet
  Kohli}.} \bibinfo{year}{2019}\natexlab{}.
\newblock \showarticletitle{Analysing mathematical reasoning abilities of
  neural models}.
\newblock \bibinfo{journal}{\emph{arXiv preprint arXiv:1904.01557}}
  (\bibinfo{year}{2019}).
\newblock


\bibitem[Snoek et~al\mbox{.}(2012)]%
        {snoek2012practical}
\bibfield{author}{\bibinfo{person}{Jasper Snoek}, \bibinfo{person}{Hugo
  Larochelle}, {and} \bibinfo{person}{Ryan~P Adams}.}
  \bibinfo{year}{2012}\natexlab{}.
\newblock \showarticletitle{Practical bayesian optimization of machine learning
  algorithms}.
\newblock \bibinfo{journal}{\emph{Advances in neural information processing
  systems}}  \bibinfo{volume}{25} (\bibinfo{year}{2012}).
\newblock


\bibitem[Stanley et~al\mbox{.}(2009)]%
        {stanley2009hypercube}
\bibfield{author}{\bibinfo{person}{Kenneth~O Stanley}, \bibinfo{person}{David~B
  D'Ambrosio}, {and} \bibinfo{person}{Jason Gauci}.}
  \bibinfo{year}{2009}\natexlab{}.
\newblock \showarticletitle{A hypercube-based encoding for evolving large-scale
  neural networks}.
\newblock \bibinfo{journal}{\emph{Artificial life}} \bibinfo{volume}{15},
  \bibinfo{number}{2} (\bibinfo{year}{2009}), \bibinfo{pages}{185--212}.
\newblock


\bibitem[Theephoowiang and Chaowicharat(2022)]%
        {theephoowiang2022difficulty}
\bibfield{author}{\bibinfo{person}{Kittipong Theephoowiang} {and}
  \bibinfo{person}{Ekawat Chaowicharat}.} \bibinfo{year}{2022}\natexlab{}.
\newblock \showarticletitle{Difficulty level estimation of mathematics problems
  using machine learning}. In \bibinfo{booktitle}{\emph{2022 4th International
  Conference on Image, Video and Signal Processing}}.
  \bibinfo{pages}{231--237}.
\newblock


\bibitem[Voevoda and Romannikov(2017)]%
        {voevoda2017synthesis}
\bibfield{author}{\bibinfo{person}{Aleksandr~Aleksandrovich Voevoda} {and}
  \bibinfo{person}{Dmitry~Olegovich Romannikov}.}
  \bibinfo{year}{2017}\natexlab{}.
\newblock \showarticletitle{Synthesis of neural network for solving
  logical-arithmetic problems}.
\newblock \bibinfo{journal}{\emph{Informatics and Automation}}
  \bibinfo{volume}{54} (\bibinfo{year}{2017}), \bibinfo{pages}{205--223}.
\newblock


\bibitem[Whitehead(2017)]%
        {whitehead2017introduction}
\bibfield{author}{\bibinfo{person}{Alfred~North Whitehead}.}
  \bibinfo{year}{2017}\natexlab{}.
\newblock \bibinfo{booktitle}{\emph{An introduction to mathematics}}.
\newblock \bibinfo{publisher}{Courier Dover Publications}.
\newblock


\bibitem[Wierstra et~al\mbox{.}(2005)]%
        {wierstra2005modeling}
\bibfield{author}{\bibinfo{person}{Daan Wierstra}, \bibinfo{person}{Faustino~J
  Gomez}, {and} \bibinfo{person}{J{\"u}rgen Schmidhuber}.}
  \bibinfo{year}{2005}\natexlab{}.
\newblock \showarticletitle{Modeling systems with internal state using
  evolino}. In \bibinfo{booktitle}{\emph{Proceedings of the 7th annual
  conference on Genetic and evolutionary computation}}.
  \bibinfo{pages}{1795--1802}.
\newblock


\end{thebibliography}


\end{document}